\documentclass{styles/svproc}

\usepackage{url}

\usepackage{multirow}
\usepackage{amsfonts}
\usepackage{layout}
\usepackage{amsmath, bm}
\usepackage{amsthm}
\usepackage{amssymb} 
\usepackage{color}
\usepackage{algorithmicx}
\usepackage{algpseudocode}
\usepackage{algorithm}
\usepackage{verbatim}
 \usepackage{tikz}
\usetikzlibrary{decorations.pathreplacing,calc}
\usepackage{enumitem}

\usepackage{epsfig}
\usepackage{caption}
\usepackage{graphicx}
\usepackage{epstopdf}
\usepackage{stfloats}

\usepackage{hyperref}
\usepackage{url}
\usepackage{breakurl}

\newcommand{\stepsize}{h}
\newcommand{\decrease}{\rho}
\newcommand{\primal}{F}
 
\newcommand{\x}{{w}}   
\newcommand{\y}{{\alpha}}   

\newcommand{\setn}{[n]}
\newcommand{\setd}{[d]}

\newcommand{\Xcal}{\mathcal W}

\newcommand\tagthis{\addtocounter{equation}{1}\tag{\theequation}}

\newcommand{\eqdef}{\stackrel{\text{def}}{=}}

\newcommand{\R}{\mathbb{R}}

\DeclareMathOperator{\proj}{proj}         

\DeclareMathOperator{\Exp}{\mathbf{E}}           

\DeclareMathOperator{\dom}{dom}         

\newtheorem{assumption}{Assumption}

\theoremstyle{plain}

\newtheorem{cor}[theorem]{Corollary}

\theoremstyle{definition}

\usepackage[colorinlistoftodos,bordercolor=orange,backgroundcolor=orange!20,linecolor=orange,textsize=scriptsize]{todonotes}

\begin{document}
\mainmatter              
\title{Projected Semi-Stochastic Gradient Descent Method with Mini-Batch Scheme under Weak Strong Convexity Assumption}
\titlerunning{Projected Semi-Stochastic Gradient Descent (PS2GD)}  
%
\author{Jie Liu \and Martin Tak\'a\v{c}} 
\authorrunning{Liu \and Tak\'a\v{c}} 
%
\tocauthor{Ivar Ekeland, Roger Temam, Jeffrey Dean, David Grove,
Craig Chambers, Kim B. Bruce, and Elisa Bertino}
\institute{Lehigh University, Bethlehem, PA 18015, USA,\\
\email{jie.liu.2018@gmail.com},\quad\email{takac.mt@gmail.com}}

\maketitle              

\begin{abstract}
We propose a projected semi-stochastic gradient descent method with mini-batch for improving both the theoretical complexity and practical performance of the general stochastic gradient descent method (SGD). We are able to prove linear convergence under weak strong convexity assumption. This requires no strong convexity assumption for minimizing the sum of smooth convex functions subject to a compact polyhedral set, which remains popular across machine learning community. Our PS2GD preserves the low-cost per iteration and high optimization accuracy via stochastic gradient variance-reduced technique, and admits a simple parallel implementation with mini-batches. Moreover, PS2GD is also applicable to dual problem of SVM with hinge loss.
\end{abstract}
%

\section{Introduction}
The problem we are interested in is to minimize a constrained convex problem,
\begin{equation}\label{prob:primal}
\min_{\x\in\Xcal} \left\{ F(\x) := g(A\x) + q^T\x \right\}.
\end{equation}
where $\x\in\Xcal\subseteq \R^d, A\in\R^{n\times d}$, and assume that $F$ can be further written as
\begin{equation}\label{prob:primal2}
F(\x) := \frac{1}{n} \sum_{i=1}^n f_i(\x).
\end{equation}

This type of problem is prevalent through machine learning community. Specifically, applications which benefit from efficiently solving this kind of problems include face detection, fingerprint detection, fraud detection for banking systems, image processing, medical image recognition, and self-driving cars etc. To exploit the problem, we further make the following assumptions:

\begin{assumption}\label{ass1}
The functions  $f_i:\R^d \to \R$ are convex, differentiable and have Lipschitz continuous gradients with constant $L > 0$. That is, 
\begin{equation*}
\| \nabla f_i(\x_1)-\nabla f_i(\x_2) \|\leq L \|\x_1-\x_2\|,
\end{equation*}
for all $\x_1,\x_2\in \R^d$, where $\|\cdot\|$ is L2 norm.
\end{assumption}

\begin{assumption}\label{ass2}
 The function $g: \R^n\to \R $ is continuously differentiable and strongly convex with parameter $\mu>0$ on its effective domain that is assumed to be open and non-empty, i.e.,  $\forall z_1, z_2 \in \dom(g)\subseteq\R^n$,
\begin{equation}\label{strongconv}
g(z_1) \geq g(z_2) + \nabla g(z_2)^T(z_1-z_2) + \frac\mu2 \|z_1-z_2\|^2.
\end{equation}
\end{assumption}

\begin{assumption}\label{ass3}
 The constraint set is a compact polyhedral set, i.e.,
\begin{equation}\label{eqn:ass3}
\Xcal = \{\x\in\R^d: C\x \leq c\},\text{ where }C\in\R^{m\times d}, c\in\R^m.
\end{equation}
\end{assumption}

\vspace{0.2em}

\begin{remark}\label{remark1}
Problem \eqref{prob:primal} usually appears in machine learning problems, where $A$ is usually constructed by a sequence of training examples $\{a_i\}_{i=1}^n \subseteq\R^d$. Note that $n$ is the number of data points and $d$ is the number of features. Problem \eqref{prob:primal2} arises as a special form of problem \eqref{prob:primal} which is a general form in a finite sum structure, which covers empirical risk minimization problems. As indicated in the problem setting, there are two formulations of the problem with different pairs of $A$ and $\Xcal$ given a sequence of labeled training examples $\{(a_i, b_i)\}_{i=1}^n$ where $a_i\in\R^d, b_i\in\R$. Define the set $[m] \eqdef \{1,2,\dots,m\}$ for any positive integer $m$.

\paragraph{\textbf{Type I Primal Setting}} A commonly recognized structure for this type of problem is to apply \eqref{prob:primal} to primal problem of finite sum structured problems and to represent $g$ as $g(A\x) = \frac1n \sum_{i=1}^n g_i(a_i^T \x)$ where $g_i$ are $\R\to\R$. In this way, $f_i$ in \eqref{prob:primal} can be defined as $f_i(\x) \eqdef g_i(a_i^T\x) + q^T\x$. We need $g_i$ to have Lipschitz continuous gradients with constants $L/\|a_i\|^2$ to fulfill Assumption \ref{ass1}, i.e.,
\begin{align*}
\| \nabla f_i(\x_1)-&\nabla f_i(\x_2) \| 
\\
&= \| (a_i\nabla g_i(a_i^T\x_1) + q)-(a_i\nabla g_i(a_i^T\x_2)+q) \|
\\
&=\|a_i\| \| \nabla g_i(a_i^T\x_1) - \nabla g_i(a_i^T\x_2) \| 
\leq \|a_i\| (L/\|a_i\|^2) \|a_i^T\x_1 - a_i^T\x_2 \|
\\
&= (L/\|a_i\|) \|a_i^T(\x_1-\x_2)\|
\\
&\leq (L/\|a_i\|) \|a_i\| \|\x_1-\x_2\| = L\|\x_1-\x_2\|,
\end{align*}
where the last inequality follows from Cauchy Schwartz inequality.

Popular problems in this type from machine learning community are logistic regression and least-squares problems by letting $q=0$, i.e., $f_i(\x) = g_i(a_i^T\x) = \log(1+\exp(-b_ia_i^T\x))$ and $f_i(\x) = g_i(a_i^T\x) = \frac12(a_i^T\x-b_i)^2$, respectively. These problems are widely used in both regression and classification problems. Our results and analyses are also valid for any convex loss function with Lipschitz continuous gradient.

To deal with overfitting and enforce sparsity to the weights $\x$ in real problems, a widely used technique is to either add a regularized term to the minimization problem or enforce constraints to $\x$, for instance,
$$\min_{\x\in\R^d} \{ f(x) + g(x)\},$$
where $g(x)=\tfrac12\lambda\|x\|^2$ is called a regularizer with regularization parameter $\lambda.$ A well-known fact is that regularized optimization problem can be equivalent to some constrained optimization problem under proper conditions~\cite{NIPS2009_3675}, where the $\ell_2$ constrained optmization problem can be denoted as
$$\min_{\x\in\Xcal} f(x), \text{ with }\Xcal = \{\x\in\R^d: \|\x\|^2\leq \lambda\}.$$
The problem of our interest is formulated to solve constrained optimization problem. Under Assumption \ref{ass3}, several popular choices of polyhedral constraints exist, such as $\Xcal = \{\x\in\R^d: \| \x\|_1 \leq \zeta\}$ and $\Xcal = \{\x\in\R^d: \| \x\|_{\infty} \leq \zeta\}$.

\paragraph{\textbf{Type II Dual Setting}} We can also apply \eqref{prob:primal} to dual form of some special SVM problems. With the same sequence of labeled training examples $\{(a_i, b_i)\}_{i=1}^n$, let us denote $a_i\eqdef(a_{i1}, \dots, a_{id})^T\in\R^d$, then an example is the dual problem of SVM with hinge loss, which has the objective function:

\begin{equation}\label{eqn:SVMdual}
 g(A\y) = \frac12\sum_{i=1}^n\sum_{j=1}^n b_ib_j a_i^Ta_j \y_i \y_j
= \frac12 \y^TA^TA\y = \frac12\|A\y\|^2
\end{equation}

where the $i^{th}$ column of $A$ is $b_ia_i$ so that $[A^TA]_{ij} = (b_ia_i)^T(b_ja_j) = b_ib_ja_i^Ta_j$ and we should also know that $A\in \R^{d\times n}$. 

By defining $a_s^{(c)} \eqdef (b_1a_{1s}, b_2a_{2s}, \dots, b_na_{ns})^T\in \R^n, \forall s\in\setd$, then $a_s^{(c)}$ is the $s^{th}$ row vector of $A$ which is also called the feature vector. By deleting unnecessary $a_s^{(c)}$ corresponds to feature $s$, we can guarantee that $\| a_s^{(c)} \|\neq 0, \forall s\in\setd$ and easily scale $a_s^{(c)}$; so similar to Type I, Type II problem can also satisfy Assumption \ref{ass1}. Under this type, $\forall i\in\setd$, $f_i$ can be written as 
$$f_i(\y) = g_i \left((a_s^{(c)})^T\y\right) + q^T\y$$ with $$g_i \left((a_s^{(c)})^T\y\right) = \frac{d}2\|(a_s^{(c)})^T\y \|^2\text{ and }q = (-1, \dots, -1)^T\in\R^n,$$
and $F(\y) = \frac1d \sum_{i=1}^d f_i(\y)$.

The dual formulation of SVM with hinge loss is 
$$\min_{\y\in\Xcal} g(A\y) + q^T\y$$
with $g$ defined in \eqref{eqn:SVMdual}, $A\in \R^{d\times n}, q = (-1, \dots, -1)^T\in\R^n$ and $\Xcal = \{\y: \y_i\in [0, \lambda n], \forall i\in\setn\}\subset\R^n$, where $\lambda$ is regularization parameter \cite{takac2013ICML}. This problem satisfies Assumptions \ref{ass1}--\ref{ass3}, which is within our problem setting. 


\end{remark}

\vspace{0.2em}

\begin{remark}
Assumption \ref{ass2} covers a wide range of problems. Note that this is not a strong convexity assumption for the original problem $F(w)$ since the convexity of $F$ is dependent on the data $A$; nevertheless, the choice of $g$ is independent of $A$. Popular choices for $g(z)$ have been mentioned in Remark \ref{remark1}, i.e., $\frac12 \| z- b\|^2$, $\frac1n\sum_{i=1}^n \log\left(1+\exp(-b_i z_i)\right)$ in Type I and $\frac12\|z\|^2$ in Type II.

\end{remark}


\paragraph{\textbf{\emph{Related Work}}} A great number of methods have been delivered to solve problem \eqref{prob:primal} during the past years. One of the most efficient algorithms that have been extensively used is FISTA \cite{fista}. However, this is considered a full gradient algorithm, and is impractical in large-scale settings with big $n$ since $n$ gradient evaluations are needed per iteration. Two frameworks are imposed to reduce the cost per iteration---stochastic gradient algorithms \cite{pegasos,OhadTong,zhangsgd,nemirovskisgd,jaggi2014communication,takac2013ICML} and randomized coordinate descent methods \cite{nesterovRCDM,richtarik,richtarikBigData,necoara2014random,approx,shalev2013stochastic,marecek2014distributed,necoara2013distributed,richtarik2013distributed,fercoq2014fast,ji2015}. However, even under strong convexity assumption, the convergence rates in expectation is only sub-linear, while full gradient methods can achieve linear convergence rates \cite{nesterov2013,Xiao2014}. It has been widely accepted that the slow convergence in standard stochastic gradient algorithms arises from its unstable variance of the stochastic gradient estimates. To deal with this issue, various variance-reduced techniques have been applied to stochastic gradient algorithms \cite{roux2012,shalev2013accelerated,svrg,Xiao2014,s2gd,defazio2014,konecny2015mS2GD,Liu2017}. These algorithms are proved to achieve linear convergence rate under strong convexity condition, and remain low-cost in gradient evaluations per iteration. As a prior work on the related topic, Zhang et al. \cite{NIPS2013_4940} is the first analysis of stochastic variance reduced gradient method with constraints, although their convergence rate is worse than our work.

The topic whether an algorithm can achieve linear convergence without strong convexity assumptions remains desired in machine learning community. Recently, the concept of \emph{weak strong convexity property} has been proposed and developed based on \emph{Hoffman bound} \cite{hoffman1952,lin2014,ji2015,ye2014,hui2016}. In particular, Ji and Wright~\cite{ji2015} \emph{first} proposed the concept as \emph{optimally strong convexity} in March 2014\footnote{Even though the concept was {first} proposed by Liu and Wright in \cite{ji2015} as \emph{optimally strong convexity}, to emphasize it as an extended version of strong convexity, we use the term \emph{weak strong convexity} as in \cite{ye2014} throughout our paper.} . Necoara \cite{Necoara2015} established a general framework for weak non-degeneracy assumptions which cover the weak strong convexity. Karimi et al.~\cite{PL} summarizes the relaxed conditions of strongly convexity and analyses their differences and connections; meanwhile, they provide proximal versions of global error bound and weak strong convexity conditions, as well as the linear convergence of proximal gradient descent under these conditions. Hui \cite{hui2016} also provides a complete of summary on weak strong convexity, including their connections. This kind of methodology could help to improve the theoretical analyses for series of fast convergent algorithms and to apply those algorithms to a broader class of problems.

\paragraph{\textbf{\emph{Our contributions}}} In this paper, we combine the stochastic gradient variance-reduced technique and weak strong convexity property based on Hoffman bound to derive a projected semi-stochastic gradient descent method (PS2GD). This algorithm enjoys three benefits. First, PS2GD promotes the best convergence rate for solving \eqref{prob:primal} without strong convexity assumption from sub-linear convergence to linear convergence in theory. Second, stochastic gradient variance-reduced technique in PS2GD helps to maintain the low-cost per iteration of the standard stochastic gradient method. Last, PS2GD comes with a mini-batch scheme, which admits a parallel implementation, suggesting probably speedup in clocktime in an HPC environment.

Moreover, we have shown in Remark \ref{remark1} that our framework covers the dual form of SVM problem with hinge loss. Instead of applying SDCA \cite{shalev2013accelerated,shalev2013stochastic}, we can also apply PS2GD as a stochastic dual gradient method.

\section{Projected Algorithms and PS2GD}

A common approach to solve \eqref{prob:primal} is to use \emph{gradient projection methods} \cite{Calamai1987,lin2014,ye2014} by forming a sequence $\{y_{k}\}$ via 
\begin{align*}
y_{k+1} = \arg \min_{\x\in \Xcal} &\left[U_{k}(\x) \eqdef F(y_{k}) + \nabla F (y_{k})^T (\x-y_{k}) + \frac{1}{2\stepsize} \|\x-y_{k}\|^2 \right],
\end{align*} 
where $U_{k}$ is an upper bound on $\primal$  if $\stepsize>0$ is a stepsize parameter satisfying $\stepsize \leq \frac1L$. This procedure can be equivalently  written using the {\em projection operator} as follows:
\begin{equation*}
y_{k+1} = \proj_{\Xcal} (y_{k} - \stepsize \nabla F (y_{k})),
\end{equation*} 
where
\begin{equation*}
\proj_\Xcal (z) \eqdef \arg\min_{\x \in\Xcal} \{ \tfrac 12  \|\x-z\|^2\}.
\end{equation*} 

In large-scale setting, instead of updating the gradient by evaluating $n$ component gradients, it is more efficient to consider  the \emph{projected stochastic gradient descent} approach, in which the proximal operator is applied to a stochastic gradient step:
\begin{equation}\label{eqn:prox-svrg update}
y_{k+1} = \proj_{\Xcal}(y_{k} - \stepsize G_{k}),
\end{equation}
where $G_{k}$ is a stochastic estimate of the gradient $\nabla F(y_k)$. Of particular relevance to our work are the SVRG \cite{svrg,Xiao2014} and S2GD \cite{s2gd} methods where the stochastic estimate of $\nabla F(y_k)$ is of the form
\begin{equation}\label{eqn:S2GDgradient}
G_{k} = \nabla F (\x) + (\nabla f _{i}(y_{k}) - \nabla f_{i}(\x)),
\end{equation}
where  $\x$ is an ``old'' reference point for which the gradient $\nabla F(\x)$ was already computed in the past, and $i \in \setn$ is picked uniformly at random. A mini-batch version of similar form is introduced as mS2GD \cite{konecny2015mS2GD} with 
\begin{equation}\label{eqn:mS2GDgradient}
G_{k} = \nabla F (\x) + \frac1b\sum_{i\in A_{k}}(\nabla f _{i}(y_{k}) - \nabla f_{i}(\x)),
\end{equation}
where the mini-batch $A_{k}\subset\setn$ of size $b$ is chosen uniformly at random. Note that the gradient estimate \eqref{eqn:S2GDgradient} is a special case of \eqref{eqn:mS2GDgradient} with $b=1$. Notice that $G_{k}$ is an unbiased estimate of the gradient: 
\[\Exp_i [G_{k}] \overset{\eqref{eqn:mS2GDgradient}}{=} \nabla F(\x) + \frac{1}{b}\cdot\frac{b}{n}\sum_{i=1}^n (\nabla f_i(y_k) - \nabla f_i(\x)) \overset{\eqref{prob:primal2}}{=} \nabla F(y_k).\]

Methods such as SVRG \cite{svrg,Xiao2014}, S2GD \cite{s2gd} and mS2GD \cite{konecny2015mS2GD} update the points $y_k$ in an inner loop, and the reference point $x$ in an outer loop. This ensures that $G_{k}$ has low variance, which ultimately leads to extremely fast convergence.


We now describe the PS2GD method in mini-batch scheme (Algorithm~\ref{alg:mS2GD}).

\begin{algorithm}[H]
\caption{PS2GD}
\label{alg:mS2GD}
\begin{algorithmic}[1]
\State \textbf{Input:} $M$ (max \# of stochastic steps per epoch); $\stepsize>0$ (stepsize); $\x_0 \in\R^d$ (starting point); linear coefficients $q\in\R^d$;  mini-batch size $b \in \setn$ 
\For {$k=0,1, 2, \dots$}
    \State Compute and store $v_{k} \leftarrow \nabla F(\x_{k}) = \tfrac{1}{n}\sum_{i} \nabla f_i(\x_{k}) = \tfrac{1}{n}\sum_{i} a_i\nabla g_i(a_i^T \x_{k})+ q$ 
    \State Initialize the inner loop: $y_{k,0} \gets \x_{k}$
    \State Let $t_k \leftarrow t \in \{1,2,\dots,M\}$ uniformly at random
    \For {$t=0$ to $t_k-1$}
	    \State Choose mini-batch $A_{kt}\subset \setn$ of size $b$ uniformly at random 
        \State Compute a stochastic estimate of $\nabla F(y_{k,t})$: 
        \State \qquad\qquad $G_{k,t} \gets v_k + \frac{1}{b}\sum_{i\in A_{kt}}[\nabla g_{i}(a_i^Ty_{k,t}) - \nabla g_{i}(a_i^T\x_{k})]a_i $
        \State $y_{k,t+1} \gets \proj_{\Xcal}(y_{k,t} - \stepsize G_{k,t})$
     \EndFor
    \State  Set $\x_{k+1} \leftarrow y_{k,t_k}$
\EndFor
\noindent
\end{algorithmic}
\end{algorithm}

The algorithm includes both outer loops indexed by epoch counter $k$ and inner loops indexed by $t$. To begin with, the algorithm runs each epoch by evaluating $v_k$, which is the full gradient of $F$ at $\x_k$, then it proceeds to produce $t_k$ --- the number of iterations in an inner loop, where $t_k=t\in \{1,2,\dots,M\}$ is chosen uniformly at random. 

Subsequently, we run $t_k$ iterations in the inner loop --- the main part of our method (Steps 8-10). Each new iterate is given by the projected update~\eqref{eqn:prox-svrg update}; however, with the stochastic estimate of the gradient $G_{k,t}$ in~\eqref{eqn:mS2GDgradient}, which is formed by using a {\em mini-batch} $A_{kt}\subset [n]$ of size $|A_{kt}|=b$. Each inner iteration takes \emph{$\mathit{2b}$ component gradient evaluations}\footnote{It is possible to finish each iteration with only $b$ evaluations for  component gradients, namely $\{\nabla f_i(y_{k,t})\}_{i\in A_{kt}}$, at the cost of having to store $\{\nabla f_i(x_k)\}_{i\in\setn}$, which is exactly the way that SAG~\cite{roux2012} works. This speeds up the algorithm; nevertheless, it is impractical for big $n$.}.


\section{Complexity Result}
\label{others}
 
In this section, we state our main complexity results and  comment on how to optimally choose the parameters of the method. Denote $\Xcal^*\subseteq\Xcal$ as the set of optimal solutions. Then following ideas from the proof of Theorem 1 in \cite{konecny2015mS2GD}, we conclude the following theorem. In Appendix \ref{appendix:thm1}, we provide the complete proof.
 
\begin{theorem}\label{s2convergence}
Let Assumptions~\ref{ass1}, \ref{ass2} and \ref{ass3} be satisfied and let $\x_*\in\Xcal^*$ be any optimal solution to~\eqref{prob:primal}. In addition, assume that the stepsize satisfies $0<\stepsize \leq \min 
\left\{
     \frac{1}{4   L \alpha(b)},\frac1{L}
 \right\}$
 and that $M$ is sufficiently large so that 
\begin{equation}\label{s2rho}
\decrease \eqdef
\frac{  
\beta
  +
  4\mu\stepsize^2  L \alpha(b)
(      
M
+ 
1 
)  
  }
  {
 \mu 
\stepsize
\left(
1- {4 \stepsize   L \alpha(b) }
\right) M
  } < 1,
\end{equation}
where $\alpha(b) = \frac{m-b}{b (m-1)}$ and $\beta$ is some finite positive number dependent on the structure of $A$ in \eqref{prob:primal} and $C$ in \eqref{eqn:ass3} ~\footnote{We only need to prove the existence of $\beta$ and do not need to evaluate its value in practice. Lemma \ref{lemma:hoffmanbound} provides the existence of $\beta$.}. Then PS2GD  has linear convergence in expectation:
\begin{equation*}
\Exp (\primal(\x_k) - \primal(\x_*)) \leq \decrease^k (\primal(\x_0)-\primal(\x_*)).
\end{equation*}
\end{theorem}

\begin{remark}

Consider the special case of strong convexity, when $F$ is strongly convex with parameter $\mu_F$, 
\begin{equation*}
F(\x) - F(\x_*) \geq  \frac{\mu_F}{2} \| \x-\x_*\|^2,
\end{equation*}
then we 
have
\begin{equation}\label{eqn:rho_c}
 \decrease = 
\frac{ 1
  }
  {
\stepsize \mu_F
(
1
-
  4\stepsize  L \alpha(b)  
)M
  }
+
\frac{      
  4\stepsize  L \alpha(b) 
\left(      
 M
+ 
  1
\right)  
  }
  {
(
1
-
  4\stepsize   L \alpha(b)  
)M
  },
\end{equation}
which recovers the convergence rate from \cite{konecny2015mS2GD} and it is better than \cite{Xiao2014} computationally since their algorithm requires computation of an average over $M$ points, while we continue with the last point, which is computationally more efficient.

In the special case when  $b=1$ we get $\alpha(b)=1$, and the rate given by \eqref{s2rho} exactly recovers the rate achieved by VRPSG \cite{ye2014} (in the case when the Lipschitz constants of $\nabla f_i$ are all equal).

\end{remark}

From Theorem~\ref{s2convergence}, it is not difficult to conclude the following corollary, which aims to detect the effects of mini-batch on PS2GD. The proof of the corollary follows from the proof of Theorem 2 in \cite{konecny2015mS2GD}, and thus is omitted.

\begin{cor}
\label{thm:optimalM}
Fix target decrease $ \decrease_* \geq \decrease$, where $\decrease$ is given by \eqref{s2rho} and $\decrease_* \in (0, 1)$. If we consider the mini-batch size $b$ to be fixed and define the following quantity,
\begin{equation*}
\tilde \stepsize^b\ \eqdef\ \sqrt{ \beta^2\left( \frac{1+\decrease}{\decrease\mu} \right)^2 + \frac\beta{4\mu\alpha(b)L}} - \frac{\beta(1+\decrease)}{\decrease\mu},
\end{equation*}
then the choice of stepsize $\stepsize_*^b$ and size of inner loops $m_*^b$, which minimizes the work done --- the number of gradients evaluated --- while having $\decrease \leq \decrease_*$, is given by the following statements.

If $\tilde \stepsize^b \leq \frac1L$, then $\stepsize_*^b = \tilde \stepsize^b$ and
\begin{align*}
 m_*^b 
=
\frac{2\kappa}{\rho} \left\{ \left( 1+\frac1\rho \right)4\alpha(b) + \sqrt{
  \frac{4\alpha(b)}{ \kappa} +   \left( 1+\frac1\rho \right)^2[4\alpha(b)]^2}
 \right\},
 \tagthis\label{eqn:mstar_b}
\end{align*}
where $\kappa \eqdef \frac{\beta L}{\mu}$ is the condition number; otherwise, $\stepsize_*^b = \frac1L$
and
\begin{equation}\label{eq:fasfawefwafewa}
m_*^b
  = \frac{\kappa + 4 \alpha(b) }
  { \rho -4 \alpha(b)   (1+\rho)}.
\end{equation}

\end{cor}

If $m_*^b < m_*^1 / b$ for some $b>1$, then mini-batching can help us reach the target decrease $\decrease_*$ with fewer component gradient evaluations. Equation~\eqref{eqn:mstar_b} suggests that as long as the condition $\tilde \stepsize^b \leq \frac1L$ holds, $m_*^b$ is decreasing at a rate roughly faster than $1/b$. Hence, we can attain the same decrease with no more work, compared to the case when $b = 1$.

\section{Numerical Experiments}
In this section, we deliver preliminary numerical experiments to substantiate the effectiveness and efficiency of PS2GD. We experiment mainly on constrained logistic regression problems introduced in Remark~\ref{remark1} (Type I), i.e.,
\begin{equation}\label{Linfproblem}
\min_{\x\in\Xcal} \{F(\x) := \frac1n\sum_{i=1}^n  \log [1+\exp(-b_i a_i^T\x)]\}, \text{ with } \Xcal = \{\x\in\R^d: \| \x\|_{\infty} \leq \zeta\},
\end{equation}
where $\{(a_i, b_i)\}_{i=1}^n$ is a set of training data points with $a_i\in\R^d$ and $b_i\in\{+1, -1\}$ for binary classification problems.

We performed experiments on three publicly available binary classification datasets, namely \emph{rcv1,  news20}~\footnote{\emph{rcv1} and \emph{news20} are available at \burl{http://www.csie.ntu.edu.tw/~cjlin/libsvmtools/datasets/}.} and \emph{astro-ph}~\footnote{Available at \burl{http://users.cecs.anu.edu.au/~xzhang/data/}.}. In a logistic regression problem, the Lipschitz constant of function $f_i$ can be derived as $L_i = \|a_i\|^2/4$. We assume (Assumption~\ref{ass1}) the same constant $L$ for all functions since all data points can be scaled to have proper Lipschitz constants. We set the bound of the norm $\zeta = 0.1$ in our experiments. A summary of the three datasets is given in Table~\ref{table: datasets}, including the sizes $n$, dimensions $d$, their sparsity as proportion of nonzero elements and Lipschitz constants $L$.

\begin{table}[H]
 
\centering
\begin{tabular}{|c|c|c|c|c|c|}
\hline
Dataset  & n & d & Sparsity & L  \\
\hline \hline 
\emph{rcv1} & 20,242 & 47,236 & 0.1568\% & 0.2500\\
\hline 
\emph{news20} & 19,996  & 1,355,191  & 0.0336\% & 0.2500\\
\hline 
\emph{astro-ph} & 62,369  & 99,757  & 0.0767\% & 0.2500\\
\hline
\end{tabular}
\captionsetup{justification=centering,margin=0.5cm}
\caption{Summary of datasets used for experiments.}
\label{table: datasets}
\end{table}
We implemented the following prevalent algorithms. SGD, SGD+ and FISTA are only enough to demonstrate sub-linear convergence without any strong convexity assumption.
\begin{enumerate}[topsep=1.5ex,itemsep=1.5ex]
\item \textbf{PS2GD b=1}: the PS2GD algorithm without mini-batch, i.e., with mini-batch size $b=1$. Although a safe step-size is given in our theoretical analyses in Theorem~\ref{s2convergence}, we experimented with various step-sizes and used the constant step-size that gave the best performance.
\item \textbf{PS2GD b=4}: the PS2GD algorithm with mini-batch size $b=4$. We used the constant step-size that gave the best performance.
\item \textbf{SGD}: the proximal stochastic gradient descent method with the constant step-size which gave the best performance in hindsight.
\item \textbf{SGD+}: the proximal stochastic gradient descent with adaptive step-size $h=h_0/(k+1)$, where $k$ is the number of effective passes and $h_0$ is some initial constant step-size. We used $h_0$ which gave the best performance in hindsight.
\item \textbf{FISTA}: fast iterative shrinkage-thresholding algorithm proposed in~\cite{fista}. This is considered as the full gradient descent method in our experiments.
\end{enumerate}

\begin{figure*}[htbp]
   \centering
    \epsfig{file=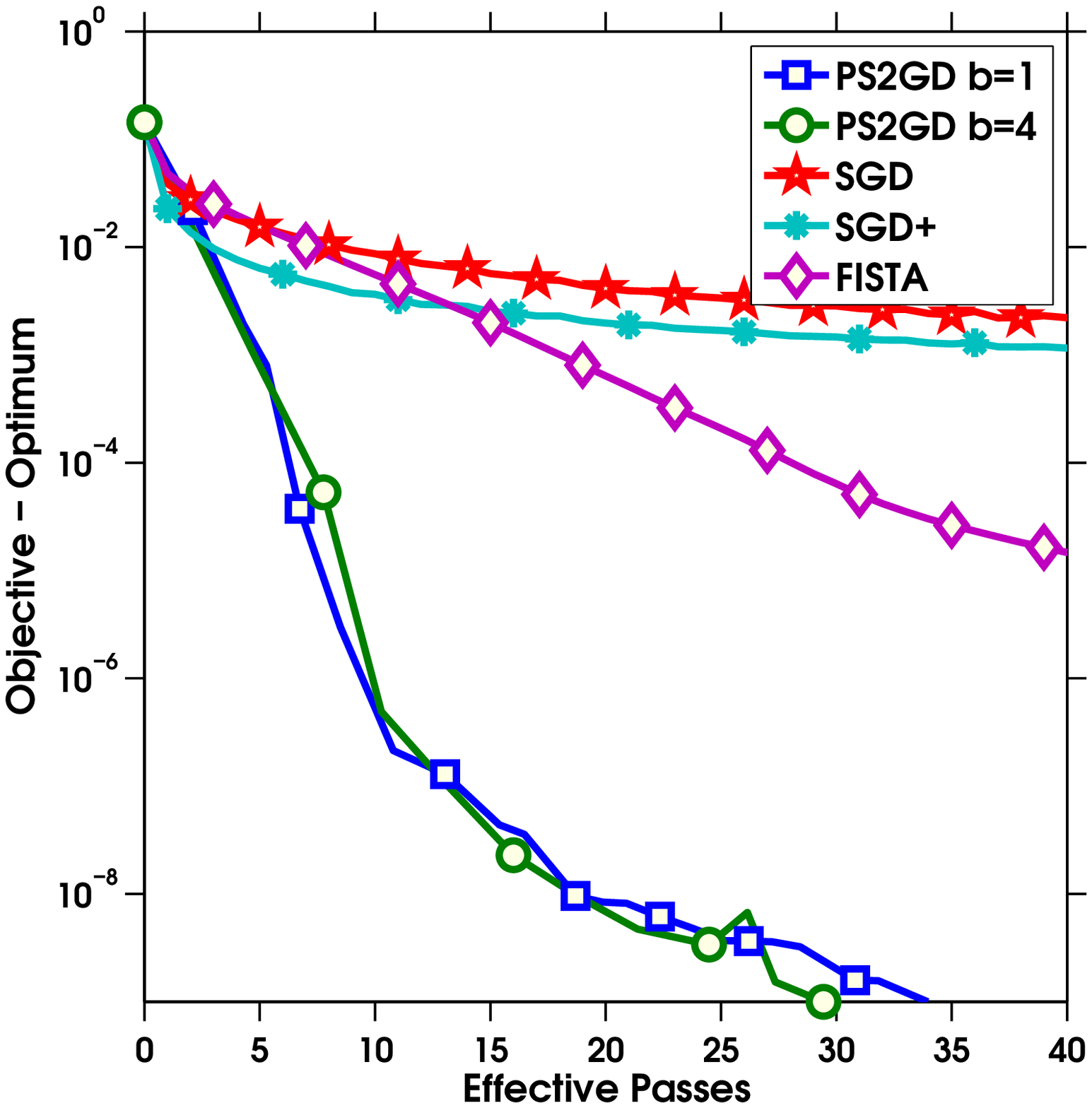,width=0.32\textwidth}
    \epsfig{file=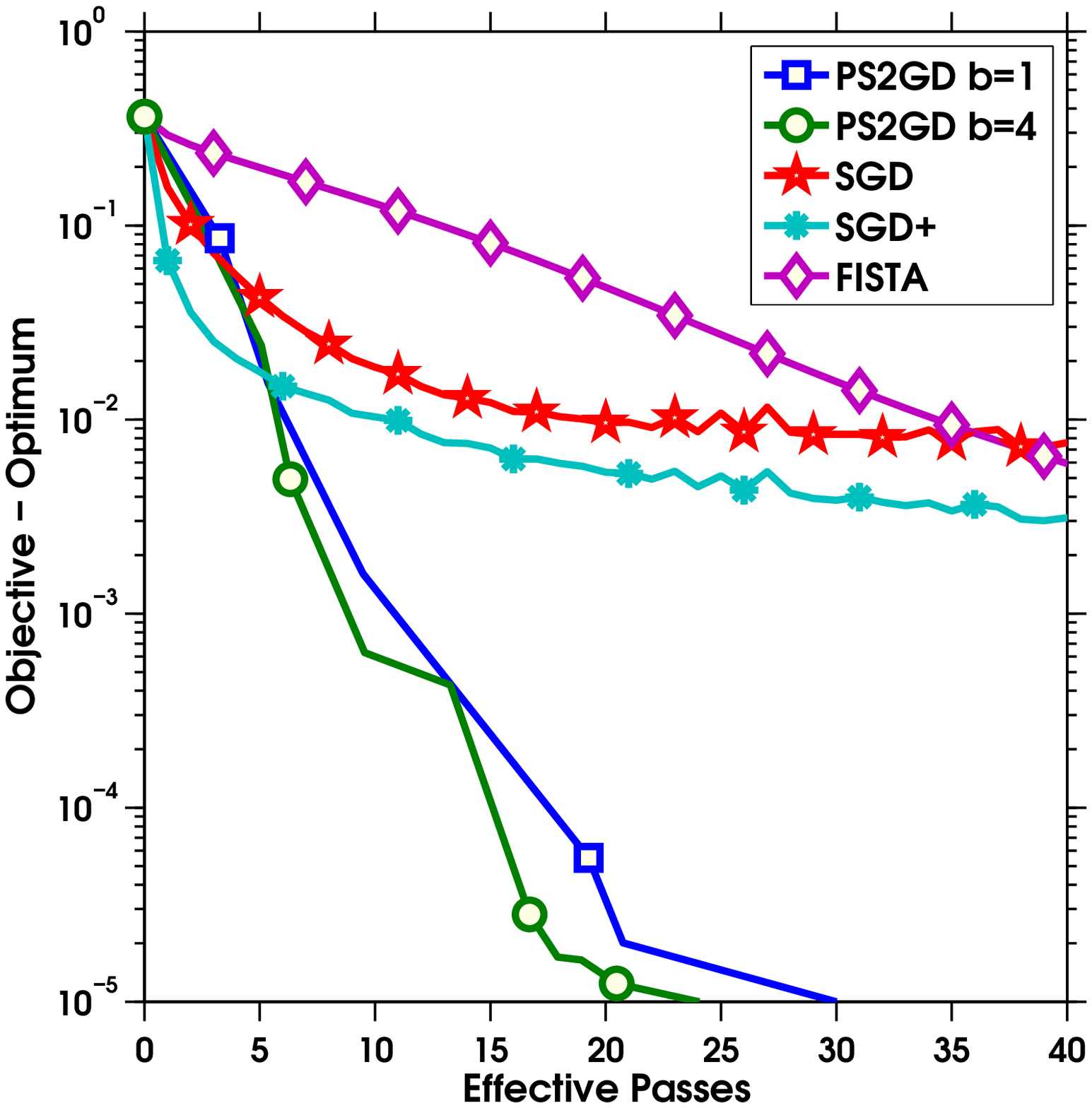,width=0.32\textwidth}
     \epsfig{file=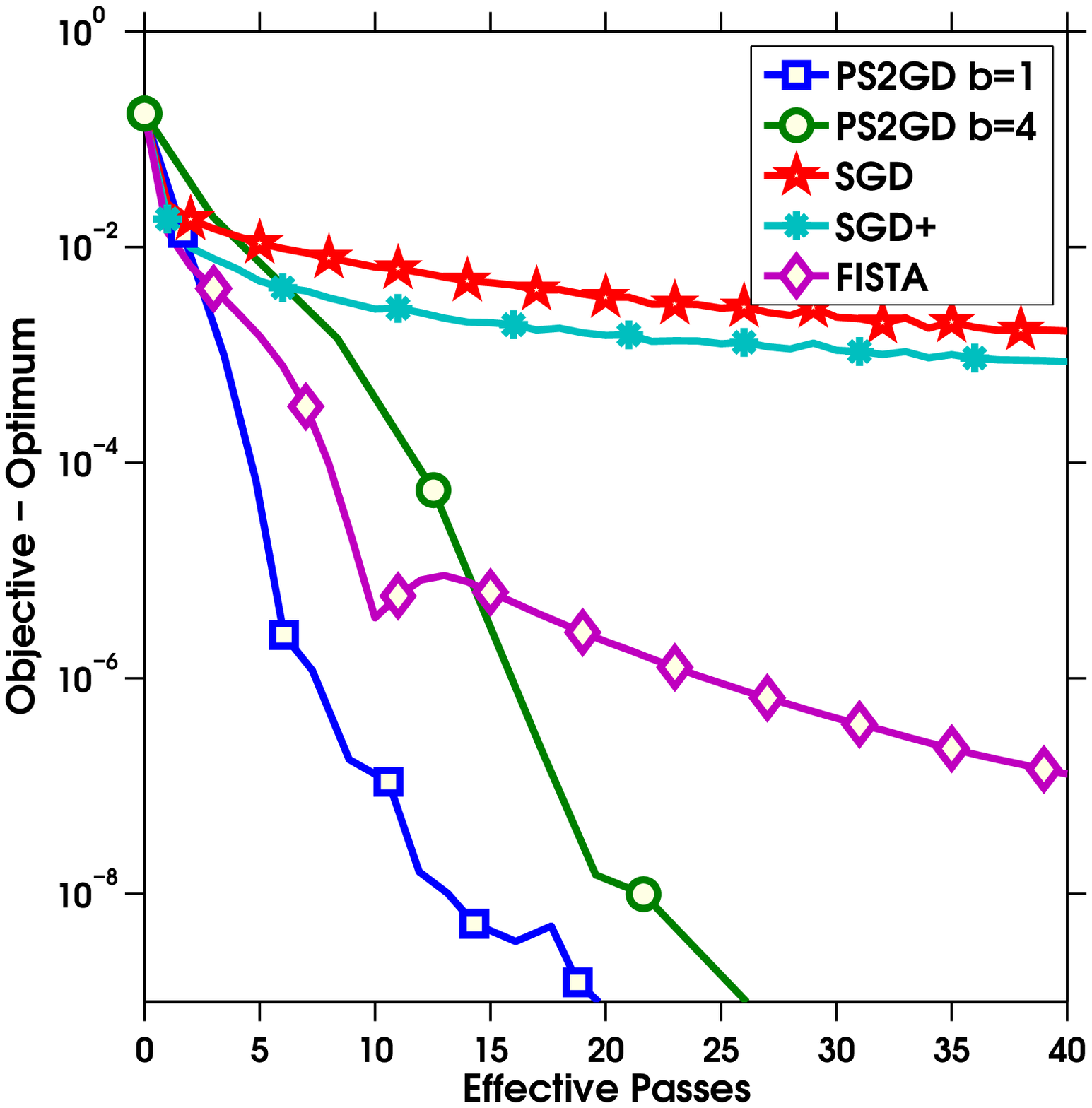,width=0.32\textwidth}
    \caption{\footnotesize Comparison of different algorithms on \emph{rcv1} (left), \emph{news20} (middle) and \emph{astro-ph}(right).}
  \label{fig:algorithms} 
\end{figure*}
In Figure~\ref{fig:algorithms}, each effective pass is considered as \emph{n} component gradient evaluations, where each $f_i$ in \eqref{prob:primal2} is named as a component function, and each full gradient evaluation counts as one effective pass.. The y-axis is the distance from the current function value to the optimum, i.e., $F(\x)-F(\x_*)$. The nature of SGD suggests unstable positive variance for stochastic gradient estimates, which induces SGD to oscillate around some threshold after a certain number of iterations with constant step-sizes. Even with decreasing step-sizes over iterations, SGD are still not able to achieve high accuracy (shown as SGD+ in Figure~\ref{fig:algorithms}). However, by incorporating a variance-reduced technique for stochastic gradient estimate, PS2GD maintains a reducing variance over iterations and can achieve higher accuracy with fewer iterations. FISTA is worse than PS2GD due to large numbers of component gradient evaluations per iteration.

\begin{figure*}[htbp]
   \centering
    \epsfig{file=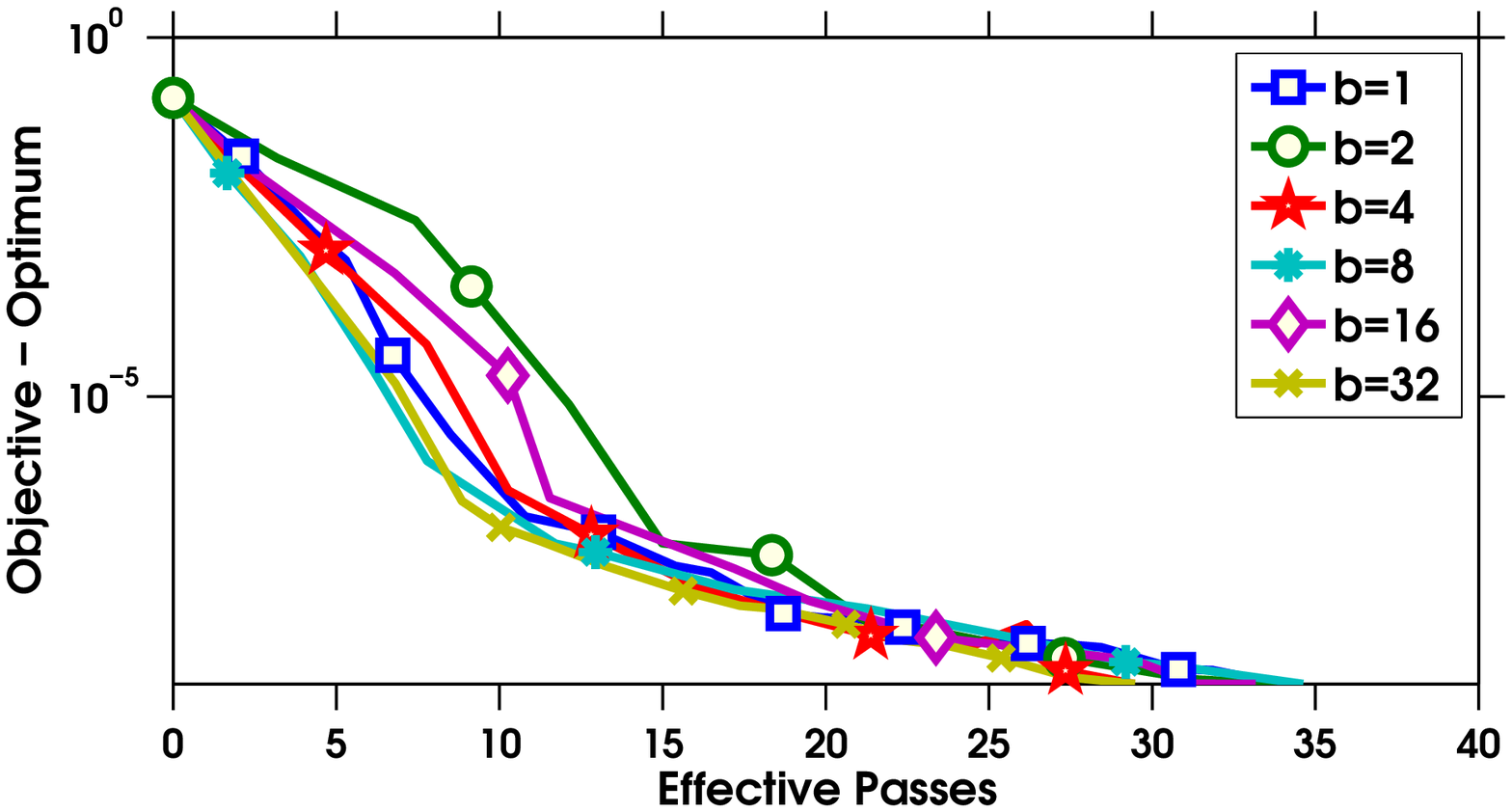,width=0.48\textwidth}
    \epsfig{file=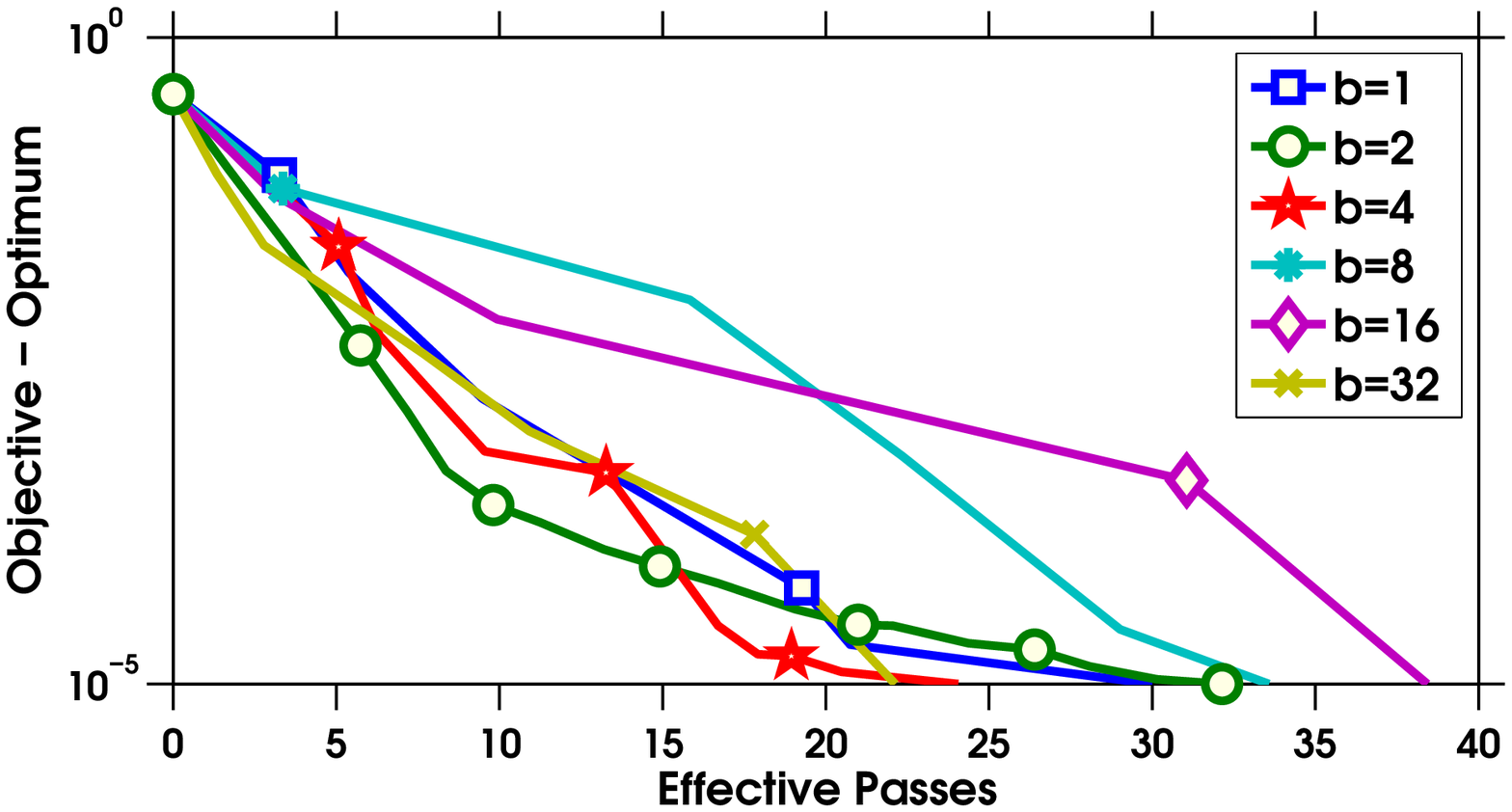,width=0.48\textwidth}
     \epsfig{file=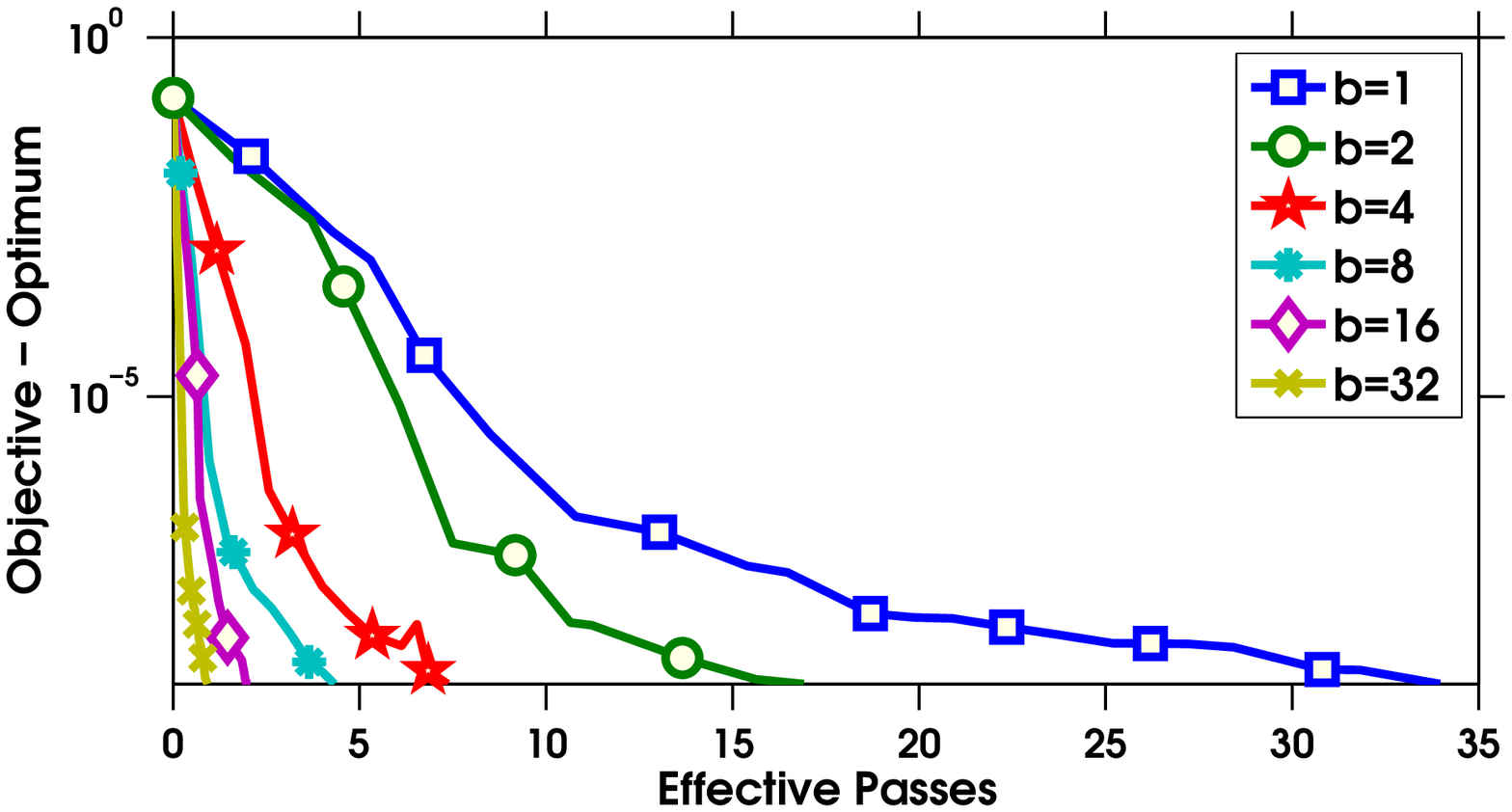,width=0.48\textwidth}
     \epsfig{file=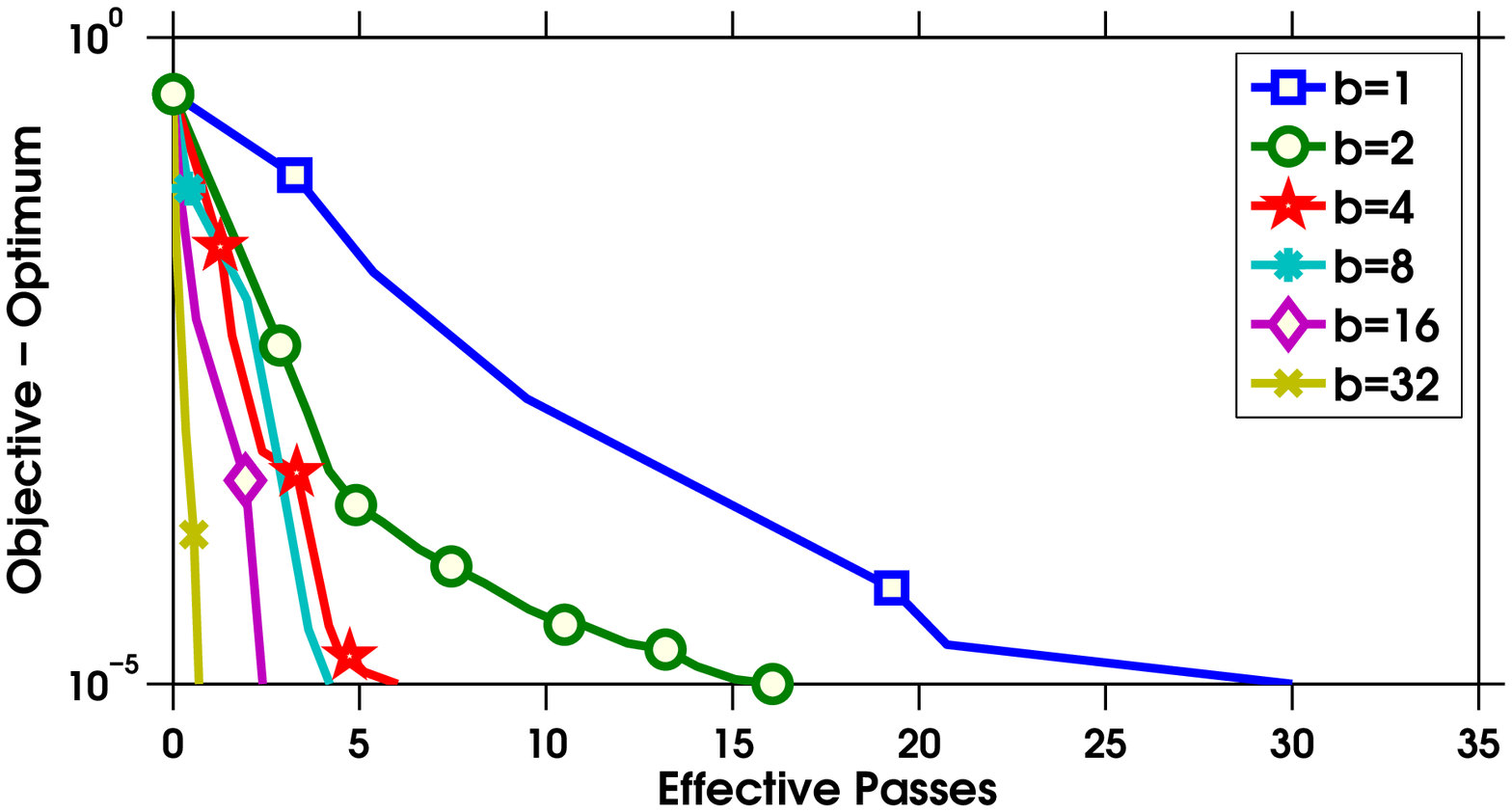,width=0.48\textwidth}
    \caption{\footnotesize Comparison of PS2GD with different mini-batch sizes on \emph{rcv1} (left) and \emph{news20} (right).}
  \label{fig:MBSpeedup} 
\end{figure*}

Meantime, increase of mini-batch size up to some threshold does not hurt the performance of PS2GD and PS2GD can be accelerated in the benefit of simple parallelism with mini-batches. Figure~\ref{fig:MBSpeedup} compares the best performances of PS2GD with different mini-batch sizes on datasets \emph{rcv1} and \emph{news20}. Numerical results on \emph{rcv1} with no parallelism imply that, PS2GD with $b=2, 4, 8, 16, 32$ are comparable or sometimes even better than PS2GD without any mini-batch ($b=1$); while on \emph{news20}, PS2GD with $b=4, 32$ are better than and the others are worse but comparable to PS2GD with $b=1$. Moreover, with parallelism, the results are promising. The bottom row shows results of ideal speedup by parallelism, which would be achievable if and only if we could always evaluate the b gradients efficiently in parallel~\footnote{In practice, it is impossible to ensure that evaluating different component gradients takes the same time; however, Figure~\ref{fig:MBSpeedup} implies the potential and advantage of applying mini-batch scheme with parallelism.}. 

\section{Conclusion}

In this paper, we have proposed a mini-batch projected semi-stochastic gradient descent method, for minimizing the sum of smooth convex functions subject to a compact polyhedral set. This kind of constrained optimization problems arise in inverse problems in signal processing and modern statistics, and is popular among the machine learning community. Our PS2GD algorithm combines the variance-reduced technique for stochastic gradient estimates and the mini-batch scheme, which ensure a high accuracy for PS2GD and speedup the algorithm. Mini-batch technique applied to PS2GD also admits a simple implementation for parallelism in HPC environment. Furthermore, in theory, PS2GD has a great improvement that it requires no strong convexity assumption of either data or objective function but maintains linear convergence; while prevalent methods under non-strongly convex assumption only achieves sub-linear convergence. PS2GD, belonging to the gradient descent algorithms, has also been shown applicable to dual problem of SVM with hinge loss, which is usually efficiently solved by dual coordinate ascent methods. Comparisons to state-of-the-art algorithms suggest PS2GD is competitive in theory and faster in practice even without parallelism. Possible implementation in parallel and adaptiveness for sparse data imply its potential in industry.

\section{Acknowledgment}
This research of Jie Liu and Martin Tak\'a\v{c} was supported by National Science Foundation grant CCF-1618717. We would like to thank Ji Liu for his helpful suggestions on related works.

\bibliographystyle{plain} 
\small{\bibliography{literature.bib}}

%
%
%
%

\clearpage


\appendix

\section{Technical Results}

\begin{lemma}\label{nonexpansiveness}
Let set $\Xcal\subseteq\R^{d}$ be nonempty, closed, and convex, then for any $x, y\in\R^d$,
\begin{equation*}
\| \proj_\Xcal (x) - \proj_\Xcal (y)\| \leq \|x-y\|.
\end{equation*}
\end{lemma}
Note that the above contractiveness of projection operator is a standard result in optimization literature. We provide proof for completeness.

Inspired by Lemma 1 in \cite{Xiao2014}, we derive the following lemma for projected algorithms.
\begin{lemma}[Modified Lemma 1 in \cite{Xiao2014}]\label{lemma:lipschitz}
Let Assumption~\ref{ass1} hold and let $\x_*\in\Xcal^*$ be any optimal solution to Problem~\eqref{prob:primal}. Then for any feasible solution $\x\in\Xcal$, the following holds,
\begin{equation}
\label{eq:varianceBound}
\frac1n\sum_{i=1}^n \| a_i [\nabla g_i (a_i^T\x) - \nabla g_i(a_i^T\x_*)] \|  = \frac1n\sum_{i=1}^n \| \nabla f_i (\x) - \nabla f_i(\x_*) \| \leq 2L[F(\x)-F(\x_*)].
\end{equation}
\end{lemma}

Lemma \ref{randvar} and Lemma \ref{lemma:hoffmanbound} come from \cite{konecny2015mS2GD} and \cite{lin2014}, respectively. Please refer to the corresponding references for complete proofs.

\begin{lemma}[Lemma 4 in \cite{konecny2015mS2GD}]\label{randvar}
Let $\{\xi_i\}_{i=1}^n$ is a collection of vectors in $\R^d$
and $\mu \eqdef \frac1n \sum_{i=1}^n \xi_i \in \R^d$.
Let $\hat S$ be a $\tau$-nice sampling. Then
\begin{equation}
\label{eq:varianceBound}
\Exp\left[ \left\|\frac1\tau \sum_{i\in \hat S} \xi_i - \mu \right\|^2 \right]
=
\frac1{n\tau}
\frac{  n-\tau}{ (n-1)}
\sum_{i=1}^n \left\| \xi_i\right\|^2.
\end{equation}

\end{lemma}

Following from the proof of Corollary 3 in \cite{Xiao2014}, by applying Lemma \ref{randvar} with $\xi_i := \nabla f _{i}(y_{k, t-1}) - \nabla f _{i}(\x_k) = a_i [\nabla g_{i}(a_i^Ty_{k, t-1}) - \nabla g_{i}(a_i^T\x_k)]$ and Lemma~\ref{lemma:lipschitz}, we have the bound for variance as follows.

\begin{theorem}
[Bounding Variance]\label{boundvariance}
Considering the definition of $G_{k, t}$ in Algorithm \ref{alg:mS2GD}, conditioned on $y_{k,t}$, we have $\Exp[G_{k, t}]=\frac{1}{n} \sum_{i=1}^n \nabla g_i(y_{k, t})+q = \nabla F (y_{k, t})$ and the variance satisfies,
\begin{align}
\Exp\left[ \|G_{k, t} - \nabla F(y_{k, t})\|^2 \right] 
 &\leq 
\underbrace{\frac{n-b}{b (n-1)} }_{\alpha(b)}
  4L [F(y_{k, t})-F(\x_*) + F(\x_k)-F(\x_*)].
  \label{eqn:bound_variance}
\end{align}
\end{theorem}

\begin{lemma}[Hoffman Bound, Lemma 15 in~\cite{lin2014}]
\label{lemma:hoffmanbound}
Consider a non-empty polyhedron
\begin{equation*}
\{ \x_* \in \R^d | C\x_*\leq c, A\x_*=r\}.
\end{equation*}
For any $\x$, there is a feasible point $\x_*$ such that
\begin{equation*}
\| \x-\x_*\| \leq \theta(A, C) 
\begin{Vmatrix}
[C\x-c]^+\\ A\x-r
\end{Vmatrix},
\end{equation*}
where $\theta(A, C)$ is independent of $x$,
\begin{equation}\label{eqn:theta}
\theta(A, C) = \sup_{u, v}
\left\{
\begin{Vmatrix}
u\\ v
\end{Vmatrix}
\left|
\begin{array}{c}
\| C^Tu+A^Tv\| = 1, u\geq 0.\text{ The corresponding rows of $C, A$ }\\ \text{to $u, v$'s non-zero elements are linearly independent. }
\end{array}
\right.
\right\}
\end{equation}
\end{lemma}

\begin{lemma}[Weak Strong Convexity]
\label{lemma:linearconvergence}
Let $\x\in\Xcal:= \{\x\in\R^d: C\x \leq c\}$ be any feasible solution (Assumption~\ref{ass3}) and $\x_* =\proj_{\Xcal^*} (\x)$ which is an optimal solution for Problem~\eqref{prob:primal}. Then under Assumptions~\ref{ass2}-\ref{ass3}, there exists a constant $\beta>0$ such that for all $\x\in\Xcal$, the following holds,
\begin{equation*}
F(\x) - F(\x_*) \geq  \frac{\mu}{2\beta} \| \x-\x_*\|^2,
\end{equation*}
where $\mu$ is defined in Assumption~\ref{ass2}. $\beta$ can be evaluated by $\beta=\theta^2$ where $\theta$ is defined in \eqref{eqn:theta}.
\end{lemma}

\section{Proofs}

\subsection{Proof of Lemma \ref{nonexpansiveness}}

For any $x, y\in\R^d$, by Projection Theorem, the following holds,
\begin{equation}\label{eqn:projection1}
[y - \proj_\Xcal (y)]^T[\proj_\Xcal(x) - \proj_\Xcal(y)]\leq 0,
\end{equation}
similarly, by symmetry, we have
\begin{equation}\label{eqn:projection2}
[x - \proj_\Xcal (x)]^T[\proj_\Xcal(y) - \proj_\Xcal(x)]\leq 0.
\end{equation}
Then \eqref{eqn:projection1} + \eqref{eqn:projection2} gives
\begin{equation*}
[\left(\proj_\Xcal(x) - \proj_\Xcal(y)\right)-(x-y)]^T[\proj_\Xcal(x)-\proj_\Xcal(y)] \leq 0,
\end{equation*}
or equivalently,
\begin{equation*}
\|\proj_\Xcal(x)-\proj_\Xcal(y)\|^2 \leq (x-y)^T[\proj_\Xcal(x)-\proj_\Xcal(y)],
\end{equation*}
and by Cauchy-Schwarz inequality, we have
\begin{equation*}
\|\proj_\Xcal(y)-\proj_\Xcal(x)\| \leq \| x-y\|,
\end{equation*}
when $\proj_\Xcal(x) =  \proj_\Xcal(y)$ are distinct; in addition, when $\proj_\Xcal(x) =  \proj_\Xcal(y)$, the above inequality also holds. Hence, for any $x, y\in\R^d$,
which is the same to
\begin{equation*}
\| \proj_\Xcal (x) - \proj_\Xcal (y)\| \leq \|x-y\|.
\end{equation*}

\subsection{Proof of Lemma \ref{lemma:lipschitz}}
For any $i\in\{1, \dots, n\}$, consider the function
\begin{equation}\label{eqn:phi}
\phi_i(\x) = f_i(\x)-f_i(\x_*) - \nabla f_i(\x_*)^T(\x-\x_*),
\end{equation}
then it should be obvious that $\nabla \phi_i(\x_*) = \nabla f_i(\x_*) - \nabla f_i(\x_*) = 0$, hence $\min_{\x\in\R^{d}}\phi_i(\x) = \phi_i(\x_*)$ because of the convexity of $f_i$. By Assumption \ref{ass1} and Remark \ref{remark1}, $\nabla \phi_i(\x)$ is Lipschitz continuous with constant $L$, hence by Theorem 2.1.5 from~\cite{nesterov2004} we have
\begin{equation*}
\frac1{2L} \|\nabla\phi_i(\x) \|^2 \leq \phi_i(\x) - \min_{\x\in\R^{l}}\phi_i(\x) = \phi_i(\x) - \phi_i(\x_*) = \phi_i(\x),
\end{equation*}
which, by~\eqref{eqn:phi}, suggests that
\begin{equation*}
\|\nabla f_i(\x) - \nabla f_i(\x_*) \|^2 \leq 2L[f_i(\x) - f_i(\x_*) - \nabla f_i(\x_*)^T(\x-\x_*)].
\end{equation*}
By averaging the above equation over $i=1, \dots, n$ and using the fact that $F(\x) = \frac{1}{n} \sum_{i=1}^n f_i(\x)$, we have
\begin{equation*}
\frac1n \sum_{i=1}^n \|\nabla f_i(\x) - \nabla f_i(\x_*) \|^2 \leq 2L[F(\x) - F(\x_*) - \nabla F(\x_*)^T(\x-\x_*)],
\end{equation*}
which, together with $\nabla F(\x_*)^T(\x-\x_*)\geq 0$ indicated by the optimality of $\x_*$ for Problem~\eqref{prob:primal}, completes the proof for Lemma~\ref{lemma:lipschitz}.

\subsection{Proof of Lemma \ref{lemma:linearconvergence}}
First, we will prove by contradiction that there exists a unique $r$ such that $\Xcal^* = \{\x\in\R^d: C\x\leq c, A\x = r\}$ which is non-empty. Assume that there exist distinct $\x_1, \x_2\in\Xcal^*$ such that $A\x_1\neq A\x_2$. Let us define the optimal value to be $F^*$ which suggests that $F^* = F(\x_1) = F(\x_2)$. Moreover, convexity of function $F$ and feasible set $\Xcal$ suggests the convexity of $\Xcal^*$, then $\frac12(\x_1+\x_2) \in \Xcal^*$. Therefore,
\begin{equation}\label{eqn:Xconv}
\begin{aligned}
F^* = F\left(\frac12(\x_1 + \x_2)\right) &\overset{\eqref{prob:primal}}{=}g\left(A\frac12(\x_1 + \x_2)\right) +\frac12q^T(\x_1 + \x_2)
\\
&= g\left(\frac12A\x_1 + \frac12A\x_2\right)+\frac12q^T(\x_1 + \x_2).
\end{aligned}
\end{equation}
Strong convexity indicated in Assumption~\ref{ass2} suggests that
\begin{align*}
F^* &= \frac12(F(\x_1)+F(\x_2)) \overset{\eqref{prob:primal}}{=} \frac12 [g (A\x_1)+q^T\x_1] + \frac12 [g(A\x_2)+q^T\x_2] \\
& = \left(\frac12g(A\x_1) + \frac12 g(A\x_2)\right)+\frac12q^T(\x_1 + \x_2)
\\
&> g\left(\frac12A\x_1 + \frac12A\x_2\right) +\frac12q^T(\x_1 + \x_2) 
\overset{\eqref{eqn:Xconv}}{=} F^*,
\end{align*}
which is a contradiction, so there exists a unique $r$ such that $\Xcal^*$ can be represented by $\{\x\in\R^d: C\x\leq c, A\x = r\}$. 

For any $\x\in\Xcal = \{x\in\R^d: C\x\leq c\}, [C\x-c]^+ = 0$, then by Hoffman's bound in Lemma~\ref{lemma:hoffmanbound}, for any $\x\in\Xcal$, there exists $\x'\in\Xcal^*$ and a constant $\theta>0$ defined in~\eqref{eqn:theta}, dependent on $A$ and $C$, such that 
\begin{equation}\label{eqn:hauffman}
\| \x-\x'\| \leq \theta
\begin{Vmatrix}
[C\x-c]^+\\ A\x-r
\end{Vmatrix}
=\theta \| A\x-r\| = \theta \| A\x - A\x_*\|, \forall \x_*\in\Xcal^*.
\end{equation}
Being aware of that by choosing $\x_*=\proj_{\Xcal_*}(\x)$, we have that $\|\x - \x_*\|\leq \|\x-\x'\|$, which suggests that 
\begin{equation*}
\| \x-\x_*\| \leq \|\x-\x'\| \overset{\eqref{eqn:hauffman}}{\leq} \theta \| A\x - A\x_*\|,
\end{equation*}
or equivalently,
\begin{equation}\label{eqn:Escale}
\| A\x - A\x_*\|^2\geq \frac1\beta\| \x-\x_*\|^2, \forall \x_*\in\Xcal^*,
\end{equation}
where $\beta= \theta^2>0$.

Optimality of $\x_*$ for Problem~\eqref{prob:primal} suggests that
\begin{equation}\label{eqn:optimality}
\nabla F(\x_*)^T(\x-\x_*) \overset{\eqref{prob:primal}}{=} [A^Tg(A\x_*)+q]^T(\x-\x_*)\geq 0,
\end{equation}
then we can conclude the following,
\begin{align}\label{strongconv1}
g(A\x) \overset{~\eqref{strongconv}}{\geq} g(A\x_*) + \nabla g(A\x_*)^T(A\x-A\x_*) + \frac\mu2 \|A\x-A\x_*\|^2,
\end{align}
which, by considering $F(\x) = g(A\x)+q^T\x$ in Problem~\eqref{prob:primal}, is equivalent to
\begin{align*}
F(\x) - F(\x_*) 
&\overset{\eqref{prob:primal}}{=} g(A\x) - g(A\x_*) + q^T(\x-\x_*) \\
&\overset{\eqref{strongconv1}}{\geq}  [A^T\nabla g(A\x_*)+ q]^T(\x-\x_*) + \frac\mu2 \|A\x-A\x_*\|^2\\
&\overset{\eqref{eqn:optimality}}{\geq}\frac\mu2 \|A\x-A\x_*\|^2\\
&\overset{\eqref{eqn:Escale}}{\geq}\frac\mu{2\beta} \|\x-\x_*\|^2.\\
\end{align*}

\subsection{Proof of Theorem \ref{s2convergence}}\label{appendix:thm1}

The proof is following the steps in \cite{konecny2015mS2GD,Xiao2014}.
For convenience, let us define the stochastic gradient mapping
\begin{equation}\label{eqn:inner_update}
d_{k, t} = \frac1\stepsize(y_{k, t}- y_{k, t+1}) = \frac1\stepsize(y_{k, t} - \proj_{\Xcal}(y_{k, t}-\stepsize G_{k, t})),
\end{equation}
then the iterate update can be written as
\begin{equation*}
y_{k, t+1} = y_{k, t} - \stepsize d_{k, t}.
\end{equation*}
Let us estimate the change of $\|y_{k, t+1}-\x_*\|$. It holds that
\begin{align}\label{eqn:thm1_eqn}
\|y_{k, t+1}-\x_*\|^2 &= \|y_{k, t}-\stepsize d_{k, t} - \x_*\|^2
= \|y_{k, t}-\x_*\|^2 - 2\stepsize d_{k, t}^T(y_{k, t}-\x_*) + \stepsize^2\|d_{k, t}\|^2.
\end{align}

By the optimality condition of $y_{k, t+1} = \proj_{\Xcal} (y_{k,t} - \stepsize G_{k,t}) = \arg\min_{\x \in\Xcal} \{ \tfrac 12  \|\x-(y_{k,t} - \stepsize G_{k,t}) \|^2\}$, we have
\begin{equation*}
[y_{k, t+1}-(y_{k} - \stepsize G_{k,t}) ]^T(\x^* - y_{k, t+1}) \geq 0,
\end{equation*}
then the update $y_{k, t+1} = y_{k, t} - \stepsize d_{k, t}$ suggests that
\begin{equation}\label{eqn:optimalitycondition}
G_{k,t}^T(\x^* - y_{k,t+1}) \geq d_{k,t}^T(\x^* - y_{k,t+1}).
\end{equation}
Moreover, Lipschitz continuity of the gradient of $F$ implies that
\begin{equation}\label{eqn:Lipschitz}
F(y_{k,t}) \geq F(y_{k,t+1}) - \nabla F(y_{k,t})^T(y_{k,t+1} - y_{k,t}) - \frac{L}{2}\|y_{k,t+1}-y_{k,t}\|^2.
\end{equation}

Let us define the operator $\Delta_{k,t} = G_{k,t}- \nabla F(y_{k,t})$, so
\begin{equation}\label{eqn:delta}
\nabla F(y_{k,t}) = G_{k,t} - \Delta_{k,t} 
\end{equation}
Convexity of $F$ suggests that
\begin{align*}
&F(\x^*) 
\geq F(y_{k,t}) + \nabla F(y_{k,t})^T ( \x^* - y_{k,t})
\\
&\overset{\eqref{eqn:Lipschitz}}{\geq}
F(y_{k,t+1}) - \nabla F(y_{k,t})^T(y_{k,t+1} - y_{k,t}) - \frac{L}{2}\|y_{k,t+1}-y_{k,t}\|^2
+ \nabla F(y_{k,t})^T ( \x^* - y_{k,t})
\\
&=F(y_{k,t+1}) - \frac{L}{2}\|y_{k,t+1}-y_{k,t}\|^2
+ \nabla F(y_{k,t})^T ( \x^* - y_{k,t+1}) 
\\
&\overset{\eqref{eqn:inner_update}, \eqref{eqn:delta}}{=}F(y_{k,t+1}) - \frac{Lh^2}{2}\|d_{k,t}\|^2
+ (G_{k,t}-\Delta_{k,t})^T ( \x^* - y_{k,t+1}) 
\\
&\overset{\eqref{eqn:optimalitycondition}}{\geq}F(y_{k,t+1}) - \frac{Lh^2}{2}\|d_{k,t}\|^2
+ d_{k,t}^T ( \x^* - y_{k,t} + y_{k,t}- y_{k,t+1})-\Delta_{k,t}^T( \x^* - y_{k,t+1})
\\
&\overset{\eqref{eqn:inner_update}}{=}F(y_{k,t+1}) - \frac{Lh^2}{2}\|d_{k,t}\|^2
+ d_{k,t}^T( \x^* - y_{k,t} + hd_{k,t})  -\Delta_{k,t}^T( \x^* - y_{k,t+1})
\\
&=F(y_{k,t+1}) + \frac{h}{2}(2-Lh)\|d_{k,t}\|^2
+ d_{k,t}^T( \x^* - y_{k,t})  -\Delta_{k,t}^T( \x^* - y_{k,t+1}) 
\\
&\overset{h\leq 1/L}{\geq}F(y_{k,t+1}) + \frac{h}{2}|d_{k,t}\|^2
+ d_{k,t}^T( \x^* - y_{k,t})  -\Delta_{k,t}^T( \x^* - y_{k,t+1}),
\end{align*}
then equivalently,
\begin{equation}\label{eqn:thm1_eqn1}
-d_{k, t}^T(y_{k, t}-\x_*) + \frac\stepsize2\|d_{k, t}\|^2 {\leq} F(\x_*)-F(y_{k, t+1})  - \Delta_{k, t}^T(y_{k, t+1}-\x_*).
\end{equation}
Therefore,
\begin{align*}
\|y_{k, t+1}-\x_*\|^2 &
\overset{\eqref{eqn:thm1_eqn},\eqref{eqn:thm1_eqn1}}{\leq} \|y_{k, t}-\x_*\|^2 
+2\stepsize
\left( 
F(\x_*)-F(y_{k, t+1})  - \Delta_{k, t}^T(y_{k, t+1}-\x_*)
\right)
\\
&\quad=
\|y_{k, t}-\x_*\|^2 
- 2\stepsize\Delta_{k, t}^T(y_{k, t+1}-\x_*) - 2\stepsize[F(y_{k, t+1})-F(\x_*)]. 
\tagthis\label{eqn:thm1_eqn2}
\end{align*}

In order to bound $-\Delta_{k, t}^T(y_{k, t+1}-\x_*)$, let us define the proximal full gradient update as\footnote{Note that this quantity is never computed during the algorithm. We can use it in the analysis nevertheless.}
\begin{equation*}
\bar{y}_{k, t+1} = \proj_{\Xcal}(y_{k, t} - \stepsize\nabla F (y_{k, t})),
\end{equation*}
with which, by using Cauchy-Schwartz inequality and Lemma~\ref{nonexpansiveness}, we can conclude that
\begin{align}
- \Delta_{k, t}^T&(y_{k, t+1}-\x_*) 
=
 - \Delta_{k, t}^T(y_{k, t+1}-\bar{y}_{k, t+1}) 
 - \Delta_{k, t+1}^T(\bar{y}_{k, t+1}-\x_*)
\nonumber
\\
&= 
-  \Delta_{k, t}^T 
\left[
\proj_{\Xcal}(y_{k, t} - \stepsize G_{k, t})- \proj_{\Xcal}(y_{k, t} - \stepsize\nabla F (y_{k, t}))
\right]
- \Delta_{k, t}^T(\bar{y}_{k, t+1}-\x_*)
\nonumber
\\
&\leq 
 \|\Delta_{k, t}\| \| (y_{k, t} - \stepsize G_{k, t})-(y_{k, t}-\stepsize\nabla F (y_{k, t}))\|
- \Delta_{k, t}^T(\bar{y}_{k, t+1}-\x_*), 
\nonumber
\\
&=  \stepsize \|\Delta_{k, t}\|^2- \Delta_{k, t}^T(\bar{y}_{k, t+1}-\x_*).
\label{eqn:thm1_eqn3}
\end{align}
So we have
\begin{align*}
&\|y_{k, t+1}-\x_*\|^2
\\
&\overset{\eqref{eqn:thm1_eqn2},\eqref{eqn:thm1_eqn3}}{\leq}
\left\|
y_{k, t}-\x_*
\right\|^2 
+2\stepsize
\left(   \stepsize \|\Delta_{k, t}\|^2- \Delta_{k, t}^T(\bar{y}_{k, t+1}-\x_*) - [F(y_{k, t+1})-F(\x_*)]\right). 
\end{align*}
By taking expectation, conditioned on $y_{k, t}$\footnote{For simplicity, we omit the $\Exp[ \cdot \,|\, y_{k, t}]$ notation in further analysis} we obtain
\begin{align}\label{eqn:thm1_eqn4}
\Exp[\|y_{k, t+1}-\x_*\|^2]
&\overset{\eqref{eqn:thm1_eqn3},\eqref{eqn:thm1_eqn2}}{\leq}
\left\|
y_{k, t}-\x_*
\right\|^2 
+ 2\stepsize
\left(   \stepsize \Exp[\|\Delta_{k, t}\|^2] - \Exp[F(y_{k, t+1})-F(\x_*)]\right),
\end{align}
where we have used that
$\Exp [\Delta_{k, t}] = \Exp[G_{k, t}] - \nabla F (y_{k, t})=0$
and hence $\Exp[- \Delta_{k, t}^T(\bar{y}_{k, t+1}-\x_*)] = 0$\footnote{$\bar{y}_{k, t+1}$ is constant, conditioned on $y_{k, t}$}.
Now, if we put \eqref{eqn:bound_variance}
into \eqref{eqn:thm1_eqn4} we obtain
\begin{align}\label{eqn:thm1_eqn5}
\Exp[\|y_{k, t+1}&-\x_*\|^2]
\leq
\left\|
y_{k, t}-\x_*
\right\|^2 
\nonumber
\\
&+ 2\stepsize
\left(4 L \stepsize \alpha(b) (F(y_{k, t})-F(\x_*) + F(\x_k)-F(\x_*)) - \Exp[F(y_{k, t+1})-F(\x_*)]\right),
\end{align}
where $\alpha(b)=\frac{m-b}{b (m-1)}$.

Now, if we consider that we have just an lower-bounds 
$\nu_F\geq 0$ of the true 
strong convexity parameter $\mu_F$,
then we obtain from \eqref{eqn:thm1_eqn5} that
\begin{align*}
\Exp[\|y_{k, t+1}&-\x_*\|^2]
\leq
\left\|
y_{k, t}-\x_*
\right\|^2 
\nonumber
\\
&+ 2\stepsize
\left(4 L \stepsize \alpha(b) (F(y_{k, t})-F(\x_*) + F(\x_k)-F(\x_*)) - \Exp[F(y_{k, t+1})-F(\x_*)]\right),
\end{align*}
which, by decreasing the index $t$ by 1, is equivalent to
\begin{align}
\Exp[\|y_{k, t}&-\x_*\|^2]
+ {2\stepsize} \Exp[F(y_{k, t})-F(\x_*)]
\leq
\left\|
y_{k, t-1}-\x_*
\right\|^2
\label{eqn:thm1_eqn6}
\\
&\qquad\qquad + 8\stepsize^2  L \alpha(b)
     (F(y_{k, t-1})-F(\x_*) + F(\x_k)-F(\x_*)).
     \nonumber
\end{align}

Now, by the definition of 
$\x_k$ we have that
\begin{align*}
\Exp[F(\x_{k+1})] 
&=
\frac1{M}
\sum_{t=1}^M 
 \Exp[F(y_{k, t})]. \tagthis\label{eqn:s2exp}
\end{align*}

By summing \eqref{eqn:thm1_eqn6} multiplied by $(1-\stepsize\nu_F)^{M-t}$ for $t=1,\dots,M$, we can obtain the left hand side
\begin{align*}
LHS = \sum_{t=1}^M
 \Exp[\|y_{k, t}-\x_*\|^2] 
+ 
2\stepsize
\sum_{t=1}^M
 \Exp[F(y_{k, t})-F(\x_*)] \tagthis\label{eqn:s2LHS}
\end{align*}
and the right hand side
\begin{align*}
RHS 
&= \sum_{t=1}^M
\Exp\|y_{k, t-1}-\x_*\|^2 
+  8\stepsize^2  L \alpha(b) 
 \sum_{t=1}^M
   \Exp[F(y_{k, t-1}) - F(\x_*) + F(\x_k) - F(\x_*)]
\\
&= \sum_{t=0}^{M-1}  \Exp\|y_{k, t}-\x_*\|^2 
+ {8\stepsize^2  L \alpha(b)}
\left(
 \sum_{t=0}^{M-1}
   \Exp[P(y_{k, t}) - P(\x_*)  ]
\right)
\\
 &\qquad\qquad +  8\stepsize^2  L \alpha(b)M
 \Exp[  F(\x_k) - F(\x_*)]
\\
&\leq \sum_{t=0}^{M-1}
     \Exp\|y_{k, t}-\x_*\|^2 
+ {8\stepsize^2  L \alpha(b)}
\left(
 \sum_{t=0}^{M} 
   \Exp[F(y_{k, t}) - F(\x_*)  ]
\right)   
\\
 &\qquad\qquad  +  8M\stepsize^2  L \alpha(b)
 \Exp[  F(\x_k) - F(\x_*)].  
\tagthis\label{eqn:s2RHS}
\end{align*}
Combining \eqref{eqn:s2LHS} and \eqref{eqn:s2RHS}
and using the fact that $LHS\leq RHS$ we have
\begin{align*}
&\Exp[\|y_{k, M} - \x_*\|^2] 
+ 
2\stepsize
\sum_{t=1}^M
\Exp[F(y_{k, t})-F(\x_*)]
\\
&\leq
\Exp\|y_{k, 0}-\x_*\|^2 
  +
  8M\stepsize^2  L \alpha(b)
 \Exp[  F(\x_k) - F(\x_*)]
\\
&+  {8\stepsize^2  L \alpha(b) }
\left(
 \sum_{t=1}^{M} 
   \Exp[F(y_{k, t}) - F(\x_*)  ]
\right)
\\
&+  {8\stepsize^2  L \alpha(b) }
   \Exp[F(y_{k, 0}) - F(\x_*)  ].   
  \end{align*}  
Now, using \eqref{eqn:s2exp} we obtain
\begin{align*}
&\Exp[\|y_{k, M}-\x_*\|^2] 
+ 
2M\stepsize ( \Exp[F(\x_{k+1})] -F(\x_*))
\\
& \leq
   \Exp\|y_{k, 0}-\x_*\|^2 
  +
 8M\stepsize^2  L \alpha(b)
 \Exp[  F(\x_k) - F(\x_*)]
\\
&+ {8M \stepsize^2  L \alpha(b) }
 \left(
      \Exp[F(\x_{k+1})]  - F(\x_*)  
\right)
\\
&+  8\stepsize^2  L \alpha(b) 
   \Exp[F(y_{k, 0}) - F(\x_*)  ].   
\tagthis \label{eqn:thm1_eqn7}
  \end{align*}  
Note that all the above results hold for any optimal solution $\x_*\in\Xcal^*$; therefore, they also hold for $\x_*' = \proj_{\Xcal^*}(\x_k)$, and Lemma~\ref{lemma:linearconvergence} implies that, under weak strong convexity of $F$, i.e., $\nu_F = 0$,
\begin{equation*}
\|\x_k - \x_*'\|^2\leq\frac{2\beta}{\mu} [F(\x_k)-F(\x_*')]\tagthis\label{eqn:strongconvex}.
\end{equation*}
Considering $\Exp\|y_{k, M} - \x_*'\|^2\geq 0$, $y_{k, 0} = \x_k,$ 
and using \eqref{eqn:strongconvex},
the inequality \eqref{eqn:thm1_eqn7} with $\x_*$ replaced by $\x_*'$ 
gives us
\begin{align*}
2 M \stepsize
\left\{
1
-
{4\stepsize   L \alpha(b) }
\right\}
&[\Exp[F(\x_{k+1})] - F(\x_*')]
\\
&\leq
\left\{
    \frac{2\beta}{\mu} 
  +
  8M\stepsize^2  L \alpha(b)
+ 8\stepsize^2  L \alpha(b)
\right\}
[F(\x_k)-F(\x_*')],
\end{align*} 
or equivalently,
\begin{equation*}
\Exp[F(\x_{k+1})-F(\x_*')]\leq \decrease[F(\x_k) - F(\x_*')],
\end{equation*}
when $1
-
 {4\stepsize   L \alpha(b) }>0$
(which is equivalent to
$ \stepsize  \leq
\frac{1}{4   L \alpha(b)}   $
), and when $\decrease$ is defined as
\begin{equation*}
\decrease = 
\frac{  
{\beta}/{\mu}
  +
  4\stepsize^2  L \alpha(b)
(      
M
+ 
1 
)  
  }
  {
\stepsize
\left(
1-{4 \stepsize   L \alpha(b) }
\right)M
  }
\end{equation*}
The above statement, together with assumptions of $h\leq 1/L$, implies $$ 0 < \stepsize \leq \min \left\{ \frac{1}{4   L \alpha(b)},\frac1{L} \right\}.$$
Applying the above linear convergence relation recursively with chained expectations and realizing that $F(\x_*') = F(\x_*)$ for any $\x_*\in\Xcal^*$ since $\x_*, \x_*'\in\Xcal^*$, we have
\begin{equation*}
\Exp [F(\x_k)-F(\x_*)] \leq \decrease^k[F(\x_0) - F(\x_*)].
\end{equation*}

\end{document}